\documentclass[fleqn,10pt]{wlscirep}
\usepackage[utf8]{inputenc}
\usepackage[T1]{fontenc}

\title{Machine Learning-Based Prediction of ICU Mortality in Sepsis-Associated Acute Kidney Injury Patients Using MIMIC-IV Database with Validation from eICU Database}

\author[1]{Shuheng Chen}
\author[1]{Junyi Fan}
\author[2]{Elham Pishgar}
\author[3]{Kamiar Alaei}
\author[4]{Greg Placencia}
\author[1,*]{Maryam Pishgar}

\affil[1]{University of Southern California, Department of Industrial and Systems Engineering, Los Angeles, California, 90087, United States}
\affil[2]{Iran University of Medical Sciences, Colorectal Research Center, Tehran, 14535, Iran }
\affil[3]{California State University, Department of Health Science, Long Beach, California, 90840, United States}
\affil[4]{California State Polytechnic University, Department of Industrial and Manufacturing Engineering, Pomona, California, 91768, United States}
\affil[*]{pishgar@usc.edu}


\begin{abstract}
Sepsis-Associated Acute Kidney Injury (SA-AKI) leads to high mortality in intensive care. This study develops machine learning models using the Medical Information Mart for Intensive Care IV (MIMIC-IV) database to predict Intensive Care Unit (ICU) mortality in SA-AKI patients. External validation is conducted using the eICU Collaborative Research Database.

For 9,474 identified SA-AKI patients in MIMIC-IV, key features like lab results, vital signs, and comorbidities were selected using Variance Inflation Factor (VIF), Recursive Feature Elimination (RFE), and expert input, narrowing to 24 predictive variables. An Extreme Gradient Boosting (XGBoost) model was built for in-hospital mortality prediction, with hyperparameters optimized using GridSearch. Model interpretability was enhanced with SHapley Additive exPlanations (SHAP) and Local Interpretable Model-agnostic Explanations (LIME). External validation was conducted using the eICU database.

The proposed XGBoost model achieved an internal Area Under the Receiver Operating Characteristic curve (AUROC) of 0.878 (95\% Confidence Interval: 0.859–0.897). SHAP identified Sequential Organ Failure Assessment (SOFA), serum lactate, and respiratory rate as key mortality predictors. LIME highlighted serum lactate, Acute Physiology and Chronic Health Evaluation II (APACHE II) score, total urine output, and serum calcium as critical features.

The integration of advanced techniques with the XGBoost algorithm yielded a highly accurate and interpretable model for predicting SA-AKI mortality across diverse populations. It supports early identification of high-risk patients, enhancing clinical decision-making in intensive care. Future work needs to focus on enhancing adaptability, versatility, and real-world applications.
\end{abstract}
\begin{document}

\flushbottom
\maketitle
%
%
\thispagestyle{empty}


\section*{Introduction}

Sepsis is a life-threatening clinical syndrome characterized by an abnormal and disproportionate immune response to an infectious insult, leading to systemic inflammation and potential organ failure \cite{pittet1995systemic, singer2016third}, remains one of the most prevalent and fatal causes of morbidity, mortality, and adverse outcomes in critically ill patients \cite{singer2016third, angus2001epidemiology, rhodes2017surviving}. Acute kidney injury (AKI) is a common complication in sepsis, affecting approximately 40\%-50\% of patients with severe sepsis \cite{peerapornratana2019acute}. The development of AKI exacerbates treatment challenges, elevates healthcare costs, and substantially increases mortality risk \cite{doi2016role, hoste2018global, chertow2005acute}. The condition known as sepsis-associated acute kidney injury (SA-AKI) is characterized by a poor prognosis, prolonged hospital stays, and a higher burden of comorbidities compared to sepsis patients without AKI \cite{peerapornratana2019acute, flannery2021sepsis}. Due to the critical nature of the condition, accurate prognostication of SA-AKI in intensive care unit (ICU) patients is crucial. 

Notable examples of scoring systems used in ICU to assess the severity of illness include the Acute Physiology and Chronic Health Evaluation II (APACHE II) \cite{da2017clinical}, the Simplified Acute Physiology Score II (SAPS II) \cite{le1993new}, and the Sequential Organ Failure Assessment (SOFA) score \cite{vincent1996sofa}. However, these systems have inherent limitations, particularly in achieving optimal specificity and sensitivity, which may compromise their accuracy in predicting patient outcomes. \cite{hu2021prediction}.  
Emerging researches highlight the growing integration of machine learning (ML) in clinical predictive analysis, facilitating the identification of complex patterns in multivariate data sets. Advanced ML models such as XGBoost, RandomForest, CatBoost, LightGBM, Support Vector Classifier (SVC), and Logistic Regression have demonstrated substantial effectiveness. Among these methods, XGBoost has gained considerable recognition for its exceptional performance in predictive modeling, particularly in clinical applications. Its capacity to effectively handle large, complex datasets, model intricate feature interactions, and maintain robustness against overfitting has made it a valuable tool for tasks such as disease prediction and patient outcome forecasting. 
For instance, Ashrafi et al. (2024) employed XGBoost to develop a prediction model for heart failure mortality in ICU patients, achieving a test AUROC of 0.9228, significantly surpassing both prior models and those reported in existing literature \cite{ashrafi2024optimizing}. Similarly, Tian et al. (2024) developed an XGBoost-based model to predict AKI in liver cirrhosis patients, achieving an AUROC of 0.832. Their work demonstrates the efficacy of machine learning in renal complication prediction \cite{tian2024prediction}. Other machine learning models have also demonstrated high performance in clinical predictive tasks. For example, Miao et al. (2024) applied the LightGBM model to predict ICU readmissions in patients with intracerebral hemorrhage (ICH), achieving an AUROC of 0.736, outperforming other models \cite{miao2024predicting}. Furthermore, Safaei et al. (2022) developed a machine learning-based prediction model for mortality in ICU patients, with CatBoost, known for its effective handling of categorical data and reduced overfitting, emerging as the top performer, achieving an AUROC of 0.86-0.92 and demonstrating strong predictive capabilities for ARDS risk stratification \cite{safaei2022catboost}.

Li, X. et al. (2023)\cite{li2023machine} proposed an XGBoost-based predictive model for ICU mortality in SA-AKI patients, but several methodological limitations reduce its overall robustness and clinical applicability. Firstly, their feature engineering relied on only 44 variables, which, while clinically relevant, likely failed to capture the full complexity of SA-AKI. By excluding a broader range of potentially informative features, the model may overlook subtle but critical patterns, limiting its predictive power. Furthermore, their approach to handling missing data, though utilizing multiple imputation, does not include stratified imputation strategies tailored to varying levels of missingness. This uniform approach could introduce bias, particularly in variables with moderate or high levels of missing data, and may compromise the model’s reliability. In terms of hyperparameter optimization, their methods lack specificity regarding the use of exhaustive and systematic techniques, such as GridSearchCV, which are essential for identifying the optimal combination of parameters and ensuring model stability. This omission raises concerns about the potential for suboptimal configurations, which may result in decreased predictive performance. Additionally, their dataset, consisting of 8129 patients from the Medical Information Mart for Intensive Care IV (MIMIC-IV) database, is both smaller and less rigorously filtered compared to more recent studies, which utilize larger, higher-quality datasets with stricter inclusion criteria. The absence of updated population criteria and the reliance on a smaller dataset increase the likelihood of biases and reduce the generalizability of their findings. Moreover, a critical limitation of their study is the absence of external validation, which is essential for assessing model generalizability across diverse patient populations. Without validation on an independent dataset, such as the eICU Collaborative Research Database, the model's performance and robustness remain uncertain, limiting its clinical applicability and reliability in real-world settings. Collectively, these limitations highlight the need for more comprehensive feature engineering, advanced missing data handling strategies, systematic hyperparameter tuning, external validation, and larger, more representative datasets to further improve the predictive accuracy and applicability of mortality risk models in critically ill SA-AKI patients.

This study aims to address the limitations observed in prior research by developing a more robust and clinically applicable machine learning model for predicting SA-AKI mortality in ICU patients. Building on a comprehensive and refined feature set, advanced missing data handling strategies, and systematic hyperparameter optimization using GridSearchCV, the proposed model seeks to enhance prognostication accuracy and outperform both traditional scoring systems and existing machine learning models. Utilizing XGBoost with rigorous data preprocessing and interpretability analyses, this study also introduces external validation using the eICU Collaborative Research Database, ensuring the model’s robustness and generalizability across diverse patient populations. By addressing the absence of external validation in previous work and implementing transparent, reproducible modeling practices, this approach bridges existing gaps in predictive performance and enhances clinical decision-making for critically ill SA-AKI patients, facilitating more timely, accurate, and actionable interventions.

\section*{Methods}\label{Methods}

\subsection*{Data Source}\label{DataSource}
This research leverages de-identified electronic health records from patients admitted to the Beth Israel Deaconess Medical Center (BIDMC) in Boston, Massachusetts, spanning the period from 2008 to 2019. The dataset is sourced from the Medical Information Mart for Intensive Care IV (MIMIC-IV) \cite{johnson2023mimic}, a publicly accessible and extensively utilized critical care database. It includes a broad range of de-identified clinical data, such as patient demographics, vital signs, lab results, medication history, clinical notes, and patient outcomes, among other variables. The dataset has been rigorously curated to ensure patient confidentiality and compliance with the Health Insurance Portability and Accountability Act (HIPAA). The integrated structure of MIMIC-IV enables a multi-dimensional analysis of patient medical histories, supporting advanced research in critical care and medical informatics.

To enhance the generalizability and robustness of the predictive model, this study also incorporates de-identified patient data from the eICU Collaborative Research Database \cite{pollard2018eicu}, which aggregates information from intensive care units across diverse hospitals in the United States, covering the period from 2014 to 2015. The eICU database is an openly accessible resource extensively employed in critical care research, offering a rich collection of clinical data, including patient demographics, physiological measurements, laboratory findings, therapeutic interventions, medication usage, and patient outcomes. Similar to MIMIC-IV, the eICU dataset is thoroughly de-identified to ensure HIPAA compliance. By applying consistent patient selection and exclusion criteria to both datasets, this study ensures methodological rigor and comparability, enabling robust external validation of the predictive model. Leveraging the complementary nature of MIMIC-IV and eICU facilitates a comprehensive evaluation of model performance across diverse patient populations and healthcare settings, supporting advanced research in critical care and predictive analytics.

\subsection*{Study Population}\label{Study Population}

\begin{figure}[htbp]
\centering
\includegraphics[width=0.67\textwidth]{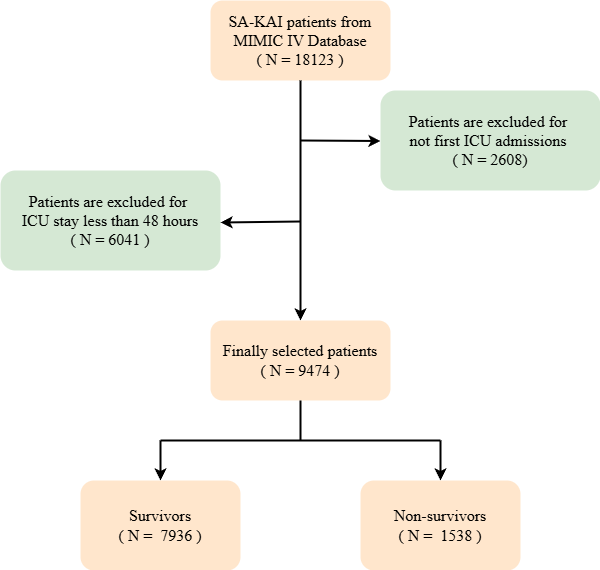}
\caption{\textbf{Study Population}}
\label{fig:study_population}
\end{figure}

The study cohort comprised patients diagnosed with sepsis who experienced AKI onset within the first 48 hours of ICU admission, identified from the MIMIC-IV database. Sepsis was defined within the first 24 hours of ICU admission according to the Sepsis-3 criteria \cite{singer2016third}, requiring a suspected infectious source and a SOFA score $\geq 2$. AKI was diagnosed using the Kidney Disease: Improving Global Outcomes (KDIGO) Clinical Practice Guidelines (2012), based on serum creatinine (Scr) levels and urine output. If no prior serum creatinine was available, the first value post-ICU admission was used as the baseline. The exclusion criteria were: (1) only the first ICU admission was included for patients with more than one ICU stays to ensure data independence, and (2) patients under 18 years of age or with ICU stays shorter than 48 hours were excluded to maintain clinical relevance. The same selection and exclusion criteria were applied to the eICU Collaborative Research Database to perform external validation and enhance generalizability. Finally, 9474 patients were indentified from MIMIC-IV database, and 8547 patients were identified from eICU Collaborative Research Database. The patient selection process for MIMIC-IV database is illustrated in \textbf{Figure~\ref{fig:study_population}.}

\subsection*{Preprocessing}\label{Preprocessing}

Data preprocessing was performed to optimize the dataset for machine learning applications. Imputation methods were applied according to the nature and proportion of missing data, ensuring accurate and reliable model inputs.

Imputation strategies for numerical variables were determined according to the proportion of missing data to maintain data quality and model robustness: (1) For variables with missing rates between 0\% and 20\%, mean imputation was applied, preserving the central tendency of the data. (2) For variables with missing rates between 20\% and 50\%, the K-Nearest Neighbors (KNN) algorithm was utilized, leveraging feature correlations to estimate missing values more accurately. (3) Variables with missing rates exceeding 50\% were excluded from the analysis to avoid potential biases and overfitting, thereby enhancing the reliability of the predictive model.

For categorical variables, imputation strategies were selected based on the proportion of missing data to ensure data integrity and model accuracy: (1) For variables with missing rates between 0\% and 20\%, the most frequent category was used for imputation, maintaining the mode of the distribution. (2) Categorical variables with missing rates exceeding 20\% were excluded from the analysis to minimize the risk of introducing noise and bias, thereby preserving the validity and reliability of the predictive model.

\subsection*{Feature Selection}\label{Feature Selection}

The feature selection process in this study was conducted through a series of well-defined stages, integrating insights from an extensive literature review, feature-selection methods including Variance Inflation Factor (VIF) and Recursive Feature Elimination (RFE), along with expert clinical input.

Initially, a comprehensive review of relevant literature was performed, and consultations with clinical experts were undertaken to validate the relevance of potential candidate features \cite{plataki2011predictors, edelstein2017biomarkers, xie2021biomarkers}. This stage resulted in a preliminary set of 96 features, which included demographic data, vital signs, comorbidities, laboratory values, and medication-related attributes, all considered relevant for the analysis. 

A three-step feature selection strategy was employed to optimize the feature set. The first step in the feature selection process involved the application of VIF to identify and exclude features with high multicollinearity, a method commonly used in contemporary research \cite{lu2021development, qu2024association, jiang2023association}. Features with a VIF value greater than 10 were removed, as they indicated significant redundancy among predictors. Following this procedure, 64 features with low VIF values were retained, thereby reducing multicollinearity and enhancing the stability and generalizability of the model. 

In the second step, RFE, a wrapper-based method, was applied to iteratively remove the least significant features, thereby identifying the most predictive variables for the target outcome. This process yielded a refined feature set comprising 21 variables, deemed most critical for accurate prediction. These features include: Length of Stay, APSIII, PaO2/FiO2 Ratio, Serum Lactate, Absolute Neutrophil Count, Lymphocyte Count, Anion Gap, Blood Glucose, Heart Rate, Serum Potassium, Partial Thromboplastin Time (PTT), Platelet Count, White Blood Cell Count (WBC), Glasgow Coma Scale (GCS), Respiratory Rate, SAPSII, SOFA, Body Temperature, Total Urine Output, Average Urine Output, and Peripheral Oxygen Saturation (SpO2). 

\begin{figure}[htbp]
\centering
\includegraphics[width=0.9\textwidth]{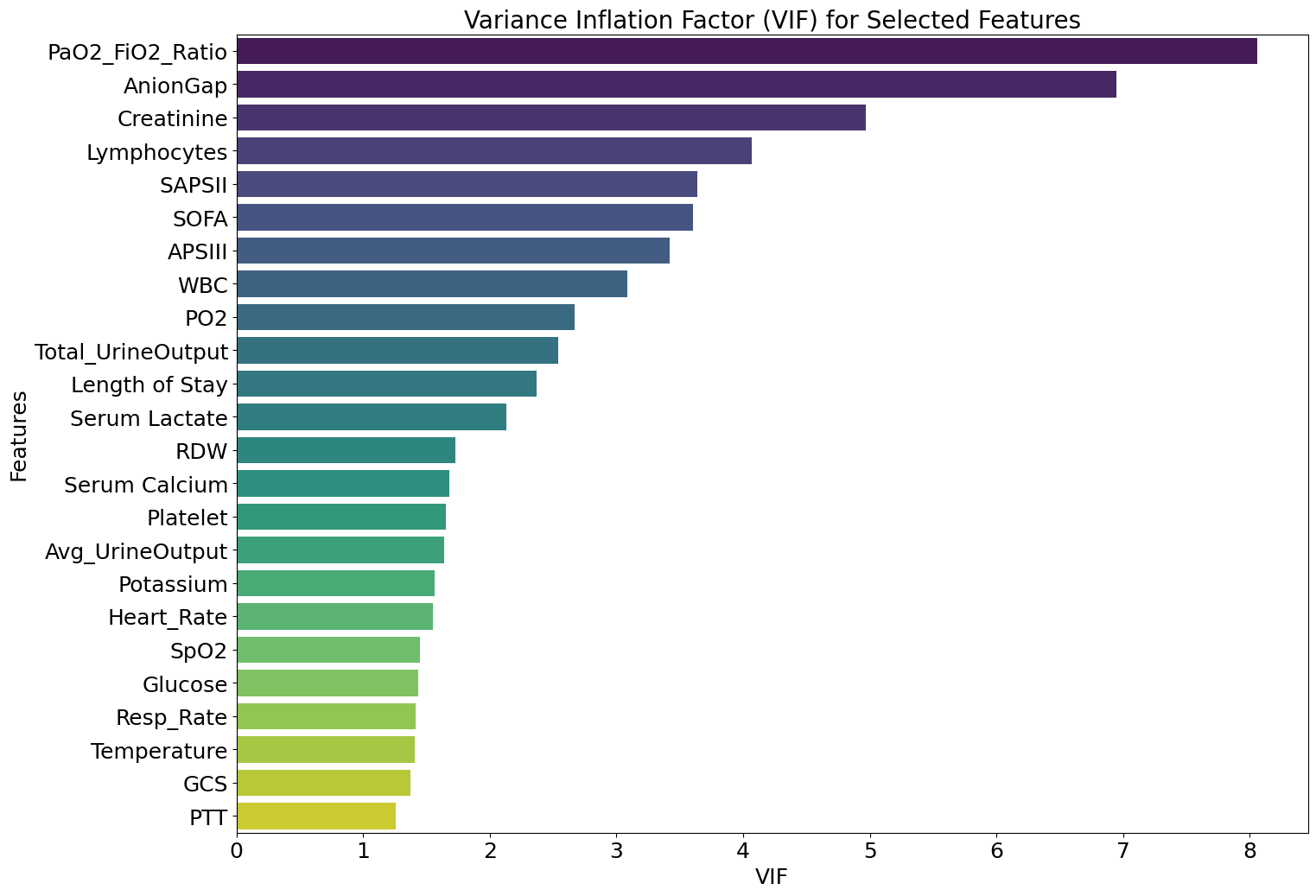}
\caption{\textbf{VIF Value of Selected Features}}
\label{fig:VIF}
\end{figure}

In the final stage, expert clinical input was incorporated to further refine the feature selection. Two clinical specialists highlighted the importance of Partial Pressure of Oxygen (PO2), Serum Calcium, and Red Cell Distribution Width (RDW), which had been excluded during the RFE process. Given their clinical relevance, these features were re-integrated into the final feature set. As a result, the refined selection consisted of 24 key predictors: Length of Stay, APSIII, PaO2/FiO2 Ratio, Serum Lactate, Absolute Neutrophil Count, Lymphocyte Count, Anion Gap, Blood Glucose, Serum Potassium, PTT, Platelet Count, WBC, GCS, SAPSII, SOFA, Total Urine Output, Average Urine Output, Heart Rate, Respiratory Rate, Body Temperature, SpO2, PO2, Serum calcium, and RDW. A detailed list of the selected features is provided in \textbf{Table~\ref{tab:clinical_features}.} And VIF value of the selected features were shown in \textbf{Figure~\ref{fig:VIF}.}

By integrating two feature-selection methods with expert clinical input, the final list of features was optimized to reflect both data-driven methods and clinical relevance. This hybrid approach improved the interpretability and robustness of the predictive model, ensuring a well-balanced set of features for subsequent analysis.

\begin{table}[htbp]
\centering
\caption{\textbf{Categorization of Selected Clinical Features}}
\label{tab:clinical_features}
\small
\begin{tabular}{l l} 
    \toprule
    \textbf{Category} & \textbf{Features} \\
    \midrule
    \textbf{Demographic and Clinical Information} & Length of Stay \\
    \midrule
    \textbf{Severity of Illness Scores} & APSIII \\
                                         & SAPSII \\
                                         & SOFA \\
                                         & GCS \\
    \midrule
    \textbf{Laboratory and Biochemical Markers} & PO2 \\
                                                & PaO2\_FiO2\_Ratio \\
                                                & Serum Calcium \\
                                                & Serum Lactate \\
                                                & AnionGap \\
                                                & Creatinine \\
                                                & Glucose \\
                                                & Potassium \\
                                                & PTT \\
                                                & Platelet \\
                                                & RDW \\
                                                & WBC \\
                                                & Lymphocytes \\
    \midrule
    \textbf{Physiological Parameters} & Total\_UrineOutput \\
                                      & Avg\_UrineOutput \\
                                      & Heart\_Rate \\
                                      & Resp\_Rate \\
                                      & Temperature \\
                                      & SpO2 \\
    \bottomrule
\end{tabular}
\end{table}

\subsection*{Statistic Analysis}\label{Statistic Analysis}

Two separate statistical analyzes were performed to assess the variations between the training and test data sets. The first aimed to determine the representativeness of the training data set relative to the test data set. The second analysis focused on distinguishing survivors from non-survivors, aiming to identify factors that may be linked to patient outcomes. Two-sided t-tests were employed exclusively to compare group means. Statistical significance was defined as a P-value of less than 0.05. This targeted strategy enabled a comprehensive evaluation of dataset attributes and revealed significant differences that might impact downstream analyses.

\subsection*{Modeling}\label{Modeling}

The performance of the proposed model was evaluated using a train-test split, which allowed 75\% of the data set to be used for training and the remaining 25\% for testing.

This method allows the model to learn from a comprehensive data set while being evaluated on unseen data, offering an accurate measure of its generalization potential. Additionally, this methodology allows for early detection of potential overfitting, ensuring that the model performs effectively on real-world data.

The data set exhibited a significant class imbalance, with only 16.2\% instances belonging to the minority class. To mitigate this issue, we employ two widely used oversampling techniques: Synthetic Minority Oversampling Technique (SMOTE) and Adaptive Synthetic Sampling (ADASYN).

SMOTE generates new synthetic samples by interpolating between a given sample of minority class and one of its \( k \)-nearest neighbors. This method has been widely applied in recent clinical machine learning studies~\cite{mohammed2020improving, gamel2024improving, ashrafi2024enhanced}. Mathematically, a synthetic sample \( x_{\text{new}} \) is created as:

\begin{equation}
x_{\text{new}} = x_i + \delta \cdot (x_{nn} - x_i), \quad \delta \sim \mathcal{U}(0,1)
\end{equation}

where \( x_i \) is a minority class instance, \( x_{nn} \) is a randomly selected neighbor from its \( k \)-nearest neighbors, and \( \delta \) is a random value drawn from the uniform distribution \( \mathcal{U}(0,1) \).

ADASYN further enhances the oversampling process by adaptively focusing on samples that are more difficult to learn. Adjusts the number of synthetic samples generated for each minority instance based on the local class distribution. This approach has also shown effectiveness in contemporary health informatic studies~\cite{majhi2023wavelet, rahmawati2024improving}. The number of synthetic samples \( G_i \) to generate for a given instance \( x_i \) is calculated as:

\begin{equation}
G_i = \frac{r_i}{\sum_{j=1}^{n} r_j} \cdot G
\end{equation}

where \( r_i \) is the estimate of the local density of the difficult samples and \( G \) is the total number of synthetic samples to generate.

The proposed model used in this study was an XGBoost, a machine learning algorithm based on gradient boosting. Known for its efficiency and accuracy, XGBoost is capable of handling both small and large datasets, capturing nonlinear relationships, and providing robust predictions. Its ability to manage missing values and support the analysis of features is particularly suitable for clinical data analysis.

Hyperparameter optimization for the model was performed using a grid search approach, which systematically explores various combinations of hyperparameter values to determine the configuration that yields the best performance for the XGBClassifier. This mechanism can be abstracted as searching in the parameter space: 

\begin{equation}
\text{Grid} = \{( \eta, \text{max\_depth}, \text{n\_estimators}) \mid \eta \in \mathcal{L}_1, \text{max\_depth} \in \mathcal{L}_2, \text{n\_estimators} \in \mathcal{L}_3 \}
\end{equation}
This tuning process significantly improved the model's predictive accuracy and operational efficiency, ensuring optimal performance for the task. 

To comprehensively evaluate the performance of the proposed model, five widely utilized machine learning algorithms—LightGBM, SVC, CatBoost, Random Forest, and Logistic Regression—were implemented as baseline models. These algorithms were selected because of their demonstrated efficiency in classification tasks and their ability to manage complex and multivariate datasets. Each model was trained and tested using the same train-test split to ensure a standardized and fair comparison. Key metrics, including predictive accuracy, precision, and sensitivity, were calculated for all models to provide a detailed assessment of their performance. This structured evaluation highlights the strengths of the proposed XGBoost model in relation to both traditional and advanced machine learning techniques. The results of the baseline models not only establish a benchmark for the proposed approach but also offer critical information on its comparative effectiveness in handling the complexities of the dataset and achieving reliable classification results.

\section*{Results}

\subsection*{Statistical Evaluation}\label{Evaluation}

T-test analysis \cite{demvsar2006statistical} comparing the training and test datasets showed that all 24 features had P-values above 0.05, with the smallest value exceeding 0.21, indicating no statistically significant differences between the two datasets. Conversely, the comparison between survivors and non-survivors demonstrated statistically significant differences in all the 24 features, with P-values approaching zero. The only exception was Creatinine, which exhibited a P-value of 0.002. This comprehensive statistical evaluation provided valuable insights into the dataset characteristics, identifying features associated with survival outcomes while confirming the representativeness of the training and test datasets. \textbf{Table~\ref{tab:feature_training_and_test}} provides a summary of the population characteristics for the training and test datasets, while the comparison between survivors and non-survivors is displayed in \textbf{Table~\ref{tab:sur and non_sur}}.

\begin{table}[htbp]
\centering
\caption{\textbf{Feature Statistics for Training and Test Sets}}
\label{tab:feature_training_and_test}
\small
\begin{tabular}{llll}
\toprule
    \textbf{Feature} & \textbf{Training Set} & \textbf{Test Set} & \textbf{P-value} \\
    \midrule
    Temperature & 36.840 [35.987--37.745] & 36.854 [36.028--37.741] & 0.212 \\
    PO2 & 113.065 [35.630--247.775] & 111.409 [35.000--245.860] & 0.221 \\
    Serum Lactate & 2.260 [0.800--6.598] & 2.211 [0.800--5.932] & 0.228 \\
    PaO2\_FiO2\_Ratio & 226.274 [75.098--427.614] & 228.760 [76.350--438.959] & 0.269 \\
    SOFA & 6.073 [1.915--12.980] & 5.998 [1.868--13.153] & 0.312 \\
    WBC & 13.147 [4.236--30.440] & 13.367 [4.399--29.597] & 0.325 \\
    Length of Stay & 6.903 [2.062--26.971] & 7.078 [2.054--27.377] & 0.376 \\
    Total\_UrineOutput & 10230.271 [63.45--48302.85] & 10498.852 [96.05--52661.45] & 0.453 \\
    SpO2 & 96.515 [92.782--99.318] & 96.547 [92.834--99.313] & 0.462 \\
    Serum Calcium & 1.124 [0.973--1.276] & 1.126 [0.977--1.284] & 0.476 \\
    Potassium & 4.270 [3.473--5.375] & 4.277 [3.454--5.400] & 0.531 \\
    APSIII & 61.096 [29.000--113.000] & 61.331 [30.000--112.000] & 0.670 \\
    Avg\_UrineOutput & 126.709 [5.696--393.348] & 127.694 [6.594--378.178] & 0.694 \\
    Resp\_Rate & 20.048 [14.064--27.795] & 20.019 [14.072--28.114] & 0.755 \\
    Creatinine & 2.581 [1.010--7.639] & 2.568 [1.008--7.282] & 0.773 \\
    Lymphocytes & 10.730 [2.000--27.000] & 10.676 [1.857--26.161] & 0.776 \\
    GCS & 14.399 [11.491--15.000] & 14.404 [12.000--15.000] & 0.820 \\
    SAPSII & 46.719 [23.000--79.000] & 46.783 [23.000--78.650] & 0.860 \\
    Platelet & 179.082 [38.000--407.360] & 178.773 [40.919--412.872] & 0.906 \\
    RDW & 16.099 [12.767--22.280] & 16.092 [12.833--22.937] & 0.910 \\
    AnionGap & 15.438 [9.683--23.911] & 15.448 [9.576--23.789] & 0.920 \\
    PTT & 41.992 [24.225--85.278] & 42.000 [24.000--87.289] & 0.986 \\
    Glucose & 146.657 [88.200--261.261] & 146.666 [89.000--251.186] & 0.993 \\
    Heart\_Rate & 85.774 [61.177--114.617] & 85.774 [61.257--115.377] & 0.999 \\
    \bottomrule
\end{tabular}
\end{table}

\begin{table}[htbp]
\centering
\caption{\textbf{Feature Statistics for Survivors and Non-survivors Sets}}
\label{tab:sur and non_sur}
\small
\begin{tabular}{llll}
\toprule
\textbf{Feature} & \textbf{Survivors} & \textbf{Non-survivors} & \textbf{P-value} \\ \midrule
Length of Stay       & 6.706 [2.059–26.913]       & 7.931 [2.078–27.185]       & 0.000 \\
Resp\_Rate           & 19.679 [14.002–26.582]     & 21.972 [14.810–30.618]     & 0.000 \\
Heart\_Rate          & 84.748 [60.966–113.159]    & 91.124 [63.428–119.572]    & 0.000 \\
Avg\_UrineOutput     & 136.456 [8.826–406.184]    & 75.908 [1.772–249.253]     & 0.000 \\
Total\_UrineOutput   & 10826.957 [100.000–51044.100]  & 7120.166 [15.000–36519.725]  & 0.000 \\
SOFA                 & 5.522 [1.871–11.322]       & 8.947 [3.207–15.788]       & 0.000 \\
SAPSII               & 45.088 [22.000–75.000]     & 55.223 [29.475–89.525]     & 0.000 \\
GCS                  & 14.456 [11.847–15.000]     & 14.101 [10.607–15.000]     & 0.000 \\
WBC                  & 12.567 [4.299–27.419]      & 16.170 [3.895–37.464]      & 0.000 \\
RDW                  & 15.904 [12.720–21.800]     & 17.115 [13.133–24.310]     & 0.000 \\
Platelet             & 182.587 [40.499–410.340]   & 160.818 [29.879–388.287]   & 0.000 \\
PTT                  & 40.330 [24.175–83.200]     & 50.658 [24.900–103.466]    & 0.000 \\
Potassium            & 4.246 [3.462–5.301]        & 4.393 [3.542–5.644]        & 0.000 \\
Glucose              & 144.891 [88.800–257.251]   & 155.858 [85.023–271.679]   & 0.000 \\
AnionGap             & 15.016 [9.549–22.500]      & 17.640 [10.948–27.302]     & 0.000 \\
Lymphocytes          & 10.969 [2.183–26.953]      & 9.485 [1.200–28.764]       & 0.000 \\
Serum Lactate        & 2.009 [0.800–4.795]        & 3.564 [0.937–11.300]       & 0.000 \\
Serum Calcium        & 1.128 [0.983–1.278]        & 1.105 [0.949–1.260]        & 0.000 \\
PaO2\_FiO2\_Ratio    & 229.696 [78.317–429.140]   & 208.439 [64.916–424.666]   & 0.000 \\
PO2                  & 114.979 [35.000–253.008]   & 103.084 [39.325–187.279]   & 0.000 \\
APSIII               & 58.301 [29.000–105.000]    & 75.661 [37.000–130.000]    & 0.000 \\
Temperature          & 36.853 [36.088–37.706]     & 36.774 [35.496–37.975]     & 0.000 \\
SpO2                 & 96.617 [93.269–99.326]     & 95.982 [90.488–99.255]     & 0.000 \\
Creatinine           & 2.580 [1.000–7.850]        & 2.584 [1.097–6.214]        & 0.002 \\ \bottomrule
\end{tabular}
\end{table}

\subsection*{Structure and Performance of Models}\label{Model Performance}

The proposed XGBoost model was configured with carefully tuned hyperparameters to optimize its performance for binary classification tasks. The objective function was set to \textbf{binary:logistic}, and the model employed \textbf{1000 estimators} with a \textbf{learning rate} of 0.025 to ensure a balance between training speed and prediction accuracy. A \textbf{maximum depth} of 7 was chosen for the trees, enabling the model to capture complex patterns in the data while minimizing the risk of overfitting.
To further enhance robustness, the model incorporated subsampling strategies, with both \textbf{subsample} and \textbf{colsample\_bytree} set to 0.8, reducing overfitting by limiting the number of samples and features used in each tree. Regularization parameters were also included to manage model complexity, with \textbf{reg\_alpha} set to 0.05 and \textbf{reg\_lambda} set to 0.08. The evaluation metric for training was \texttt{"logloss"}, focusing on optimizing probabilistic predictions.
An \textbf{early stopping} mechanism was applied during training, with a patience of 10 rounds, a minimum improvement threshold of \(10^{-4}\), and an option to save the best model. This mechanism prevented overfitting and ensured optimal model performance during validation. All experiments were conducted with a fixed \textbf{random state} of 42 to ensure reproducibility. These hyperparameter settings were fine-tuned to achieve high predictive accuracy and robustness, making the model well-suited for the classification tasks in this study.

\begin{figure}[htbp]
\centering
\includegraphics[width=1\textwidth]{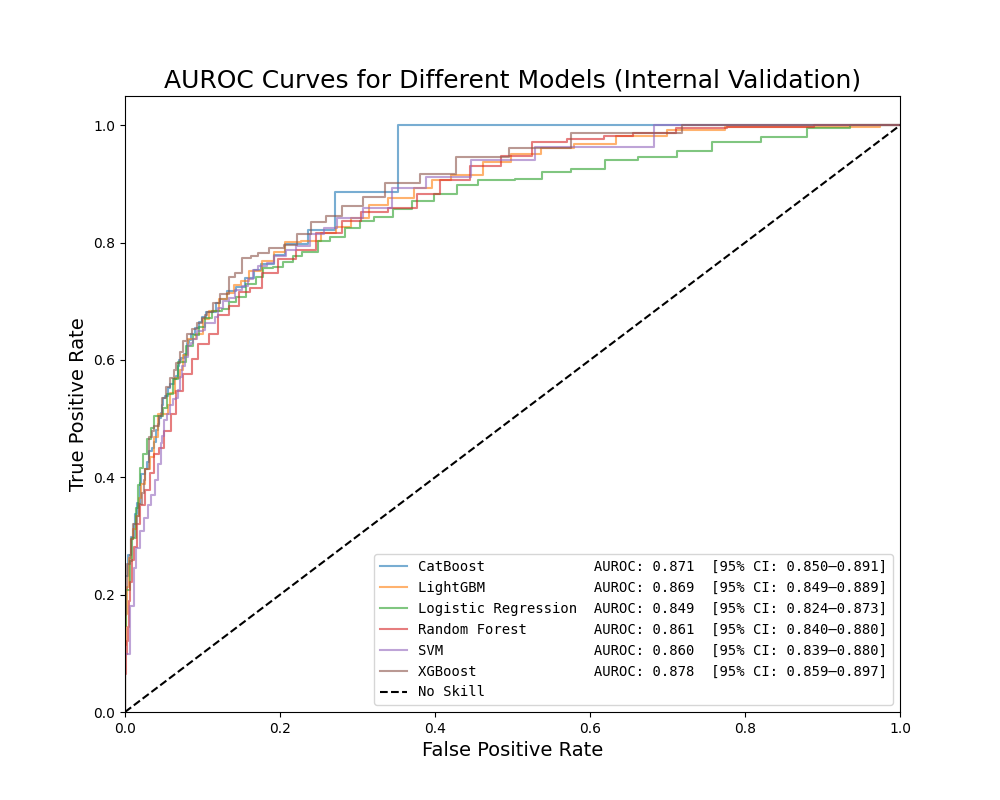}
\caption{\textbf{AUROC-Curves for Different Models (Internal Validation)}}
\label{fig:auroc}
\end{figure}

The XGBoost model achieved the highest performance among all models evaluated, with an Area Under the Receiver Operating Characteristic Curve (AUROC) of 0.878 (95\% CI: 0.859–0.897). In addition, the model demonstrated an accuracy of 0.769, a sensitivity (recall) of 0.814, and a specificity of 0.760, highlighting its robust predictive capability. Detailed performance metrics for the models are presented in \textbf{Table~\ref{tab:Performance Metrics for Models}.} The findings highlight the model's proficiency in reliably identifying high-risk patients who require closer monitoring, while achieving a well-balanced compromise between sensitivity and specificity. The model's elevated sensitivity demonstrates its effectiveness in detecting patients with a high likelihood of mortality, whereas its specificity underscores its ability to accurately recognize patients with a low probability of mortality. Together, these metrics underscore the robustness and reliability of the XGBoost model in detecting complex, non-linear patterns within the data, highlighting its efficacy as a predictive tool for mortality in SA-AKI patients. The AUROC curves for the proposed model, as well as the baseline models, are shown in \textbf{Figure~\ref{fig:auroc}.}

\begin{figure}[htbp]
\centering
\includegraphics[width=1\textwidth]{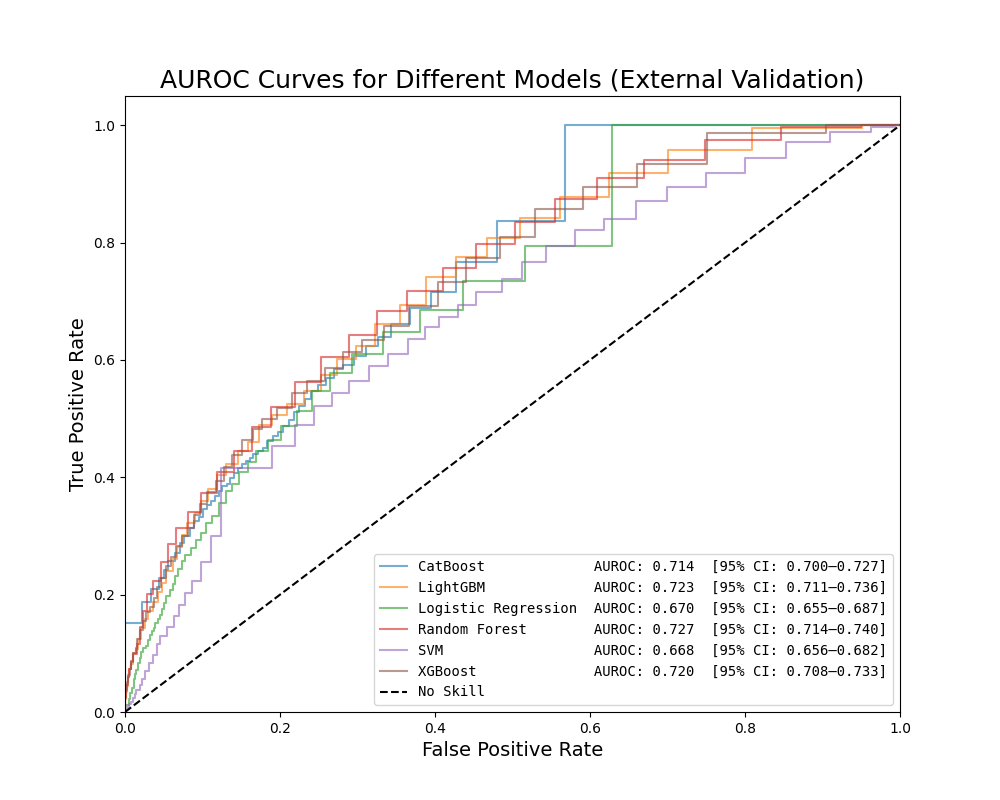}
\caption{\textbf{AUROC-Curves for Different Models (External Validation)}}
\label{fig:auroc_external}
\end{figure}

The proposed XGBoost model achieved a moderate result in external validation. Specifically, it yielded an AUROC of 0.720 (95\% CI: 0.708–0.733), an accuracy of 0.659, a sensitivity of 0.633, and a specificity of 0.665, as shown in \textbf{Table~\ref{tab:Performance Metrics for Models (external)}.} Although the model demonstrated acceptable discrimination ability, the decrease in performance compared to the internal validation indicates potential overfitting to the development dataset or variations in patient populations and clinical practices between the MIMIC-IV and eICU databases. The detailed result of external validation is presented in \textbf{Figure~\ref{fig:auroc_external}.}

\begin{table}[htbp]
\centering
\caption{Performance Metrics for Models (Internal Validation)}
\label{tab:Performance Metrics for Models}%
\small
\begin{tabular}{@{}lccccc@{}}
\toprule
\textbf{Model} & \textbf{AUROC} & \textbf{Accuracy} & \textbf{Sensitivity} & \textbf{Specificity} & \textbf{Precision} \\\midrule
\textbf{XGBoost}            & \textbf{0.878 (0.859–0.897)} & \textbf{0.769} & \textbf{0.814} & 0.760 & \textbf{0.398} \\
LightGBM           & 0.869 (0.849–0.889) & 0.757 & 0.803 & 0.748 & 0.382 \\
Random Forest      & 0.861 (0.840–0.880) & 0.759 & 0.787 & 0.754 & 0.384 \\
SVC                & 0.860 (0.839–0.880) & 0.766 & 0.795 & 0.761 & 0.393 \\
CatBoost           & 0.871 (0.850–0.891) & \textbf{0.769} & 0.798 & \textbf{0.765} & 0.397 \\
Logistic Regression & 0.849 (0.824–0.873) & 0.757 & 0.784 & 0.751 & 0.380 \\
\bottomrule
\end{tabular}
\end{table}

\begin{table}[htbp]
\centering
\caption{Performance Metrics for Models (External Validation)}
\label{tab:Performance Metrics for Models (external)}%
\small
\begin{tabular}{@{}lccccc@{}}
\toprule
\textbf{Model} & \textbf{AUROC} & \textbf{Accuracy} & \textbf{Sensitivity} & \textbf{Specificity} & \textbf{Precision} \\\midrule
XGBoost            & 0.720 (0.708–0.733) & 0.659 & 0.633 & 0.665 & 0.324 \\
LightGBM           & 0.723 (0.711–0.736) & 0.649 & 0.661 & 0.646 & 0.321 \\
Random Forest      & 0.727 (0.714–0.740) & 0.669 & 0.641 & 0.675 & 0.334 \\
SVC                & 0.668 (0.656–0.682) & 0.618 & 0.636 & 0.613 & 0.294 \\
CatBoost           & 0.714 (0.700–0.727) & 0.654 & 0.661 & 0.657 & 0.321 \\
Logistic Regression & 0.670 (0.655–0.687) & 0.625 & 0.648 & 0.625 & 0.301 \\
\bottomrule
\end{tabular}
\end{table}

\clearpage 

\clearpage

\subsection*{Interpretability}\label{Interpretability}

\subsubsection*{Feature Importance Analysis Using SHAP}\label{SHAP}
\begin{figure}[htbp]
\centering
\includegraphics[width=1\textwidth]{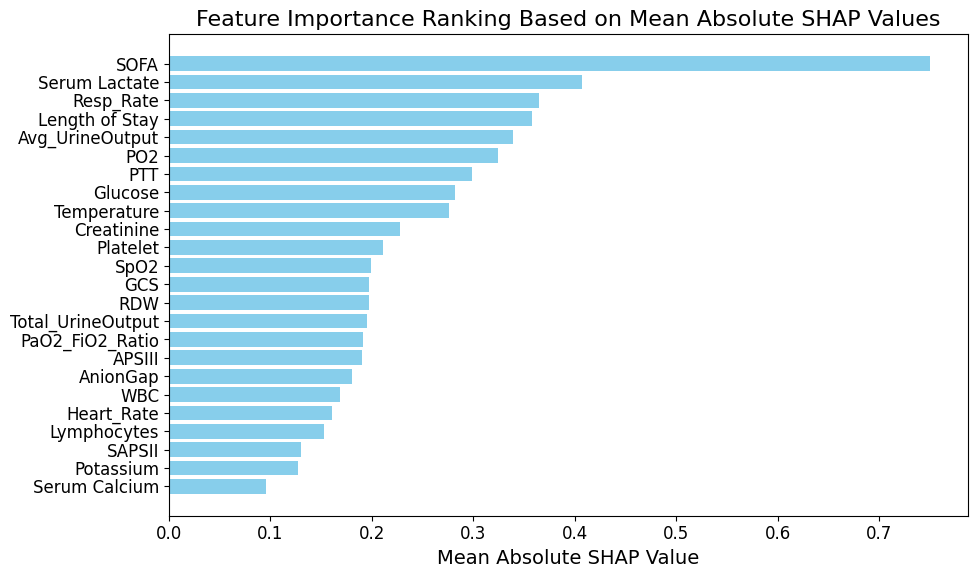}
\caption{\textbf{Feature Importance by Using SHAP}}
\label{fig:Feature Importance}
\end{figure}

To understand the model's predictions more comprehensively, Shapley Additive Explanations (SHAP) were employed, offering a powerful approach to quantify the impact of each feature on the model's outputs \cite{lundberg2017unified}, which is widely applied in recent studies \cite{hu2022explainable,li2023development,yi2023xgboost}. SHAP analysis was conducted on the test dataset, providing valuable insights into the relative importance and directional impact of features in predicting mortality among SA-AKI patients. \textbf{Figure~\ref{fig:Feature Importance}} shows the feature importance rankings calculated from mean absolute SHAP values, highlighting the SOFA score as the most influential predictor, significantly surpassing the impact of other features. It is followed by Serum Lactate, Resp\_Rate, and Length of Stay, all of which made significant contributions to the model's performance, emphasizing their critical roles in predicting mortality within the SA-AKI population.

\begin{figure}[htbp]
\centering
\includegraphics[width=0.67\textwidth]{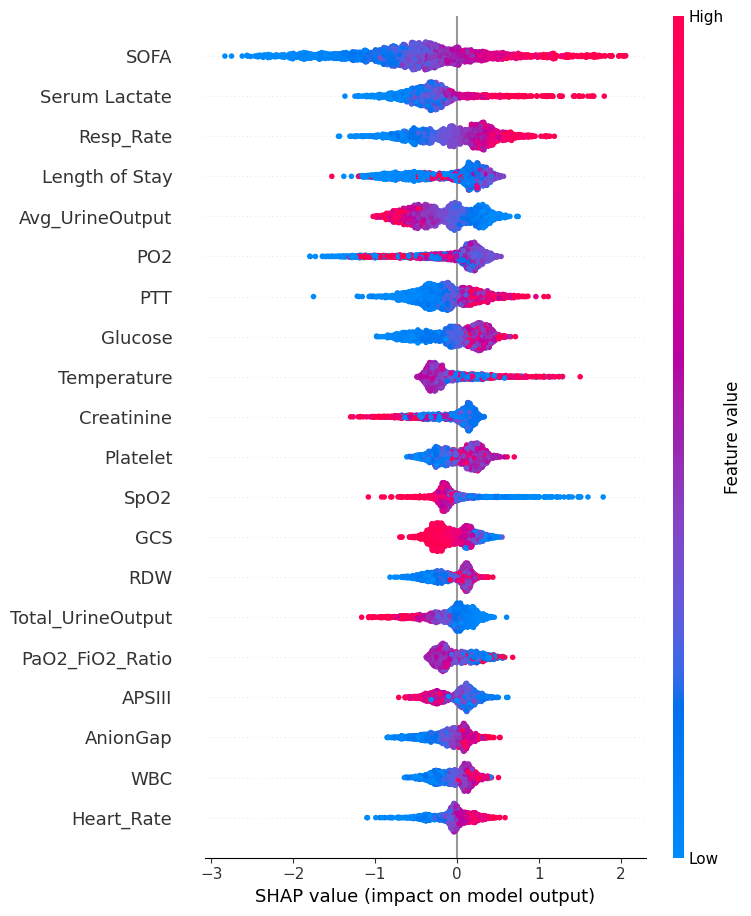}
\caption{\textbf{Feature Importance by Using SHAP}}
\label{fig:SHAP}
\end{figure}

To further examine the impact of individual feature values on the model's predictions, \textbf{Figure~\ref{fig:SHAP}} presents a SHAP summary plot, illustrating the distribution of SHAP values for all features. Each point on the plot represents an individual patient, with its color representing the feature value (low values in blue and high values in red). For instance, greater age values are linked to higher SHAP values, indicating an increased probability of readmission. Similarly, deviations in Chloride, Monocytes, and SpO2 levels appear to substantially influence the model’s predictions, either raising or lowering the estimated risk.

These findings underscore the model's ability to identify and leverage key features closely associated with mortality, effectively capturing complex, nonlinear interactions within diverse clinical and demographic datasets. By providing a holistic view of feature importance, SHAP analysis improves the model's interpretability and provides valuable insights for clinical decision-making. The identified features, especially top ones like SOFA scores, Serum Lactate, and Resp\_rate, collectively highlight critical physiological and clinical factors linked to mortality risk. This insight can aid clinicians in prioritizing high-risk patients for early interventions and optimizing resource allocation for SA-AKI patients in ICU settings, ultimately improving patient outcomes.

\subsubsection*{Local Feature Contributions to Mortality Risk Using LIME}\label{LIME}

Local Interpretable Model-agnostic Explanations (LIME) \cite{10.1145/2939672.2939778} were implemented to perform a detailed and interpretable analysis of the model’s predictions at the individual level. LIME is a widely recognized interpretability technique that explains the predictions of any machine learning model by locally approximating it with a simpler, interpretable model. This approach enables the identification of the specific contributions of input features to each individual prediction. By perturbing the input data and observing the resulting changes in the model output, LIME generates feature importance scores that highlight the most influential factors in the decision-making process. 

\begin{figure}[htbp]
\centering
\includegraphics[width=1\textwidth]{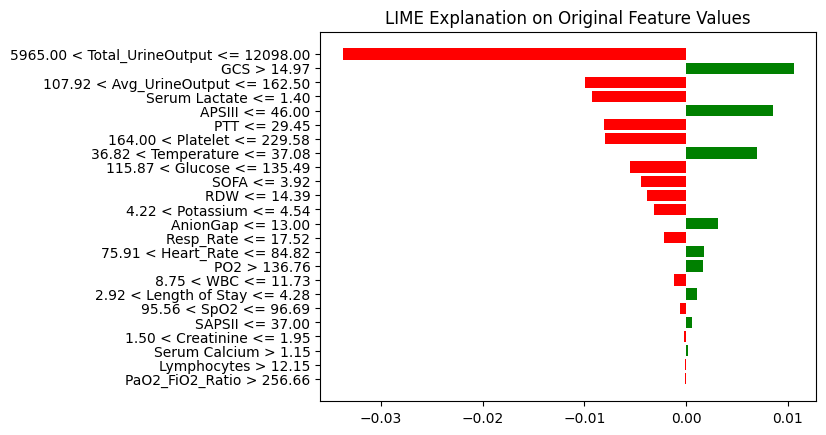}
\caption{\textbf{Local Feature Contributions to Model Prediction via LIME}}
\label{fig:LIME}
\end{figure}

As shown in \textbf{Figure~\ref{fig:LIME},} the top contributors to increased mortality risk include GCS, Serum Lactate, and APACHE II score, while total urine output, average urine output, and Serum Calcium are the most significant indicators of decreased mortality risk. The interpretability analysis highlights the nuanced role of these clinical variables, reflecting the model's capacity to discern complex interactions between features and outcomes. This transparency not only enhances the reliability of the predictions but also provides clinicians with valuable insights to guide patient management. By identifying high-risk patients through such a data-driven approach, this framework has the potential to support early interventions and improve clinical outcomes in critical care settings.

\subsection*{Comparison with Best Existing Literature}\label{Comparison}

Li et al. (2023) proposed an XGBoost-based model for predicting mortality in SA-AKI patients in the ICU. Their model achieved an AUROC of 0.794 (95\% CI: 0.762–0.827), representing the best-performing model for the same predictive task in the existing literature. The performance metrics reported for their model include an accuracy of 0.832, sensitivity of 0.793, specificity of 0.752, and precision of 0.660 \cite{li2023machine}.
\textbf{Table~\ref{tab:Comparison of Performance Metrics}} provides a detailed comparison between our proposed model and the best existing model.

\begin{table}[htbp]
\centering
\caption{\textbf{Comparison of Performance Metrics }}\label{tab:Comparison of Performance Metrics}%
\small
\begin{tabular}{@{}lccccc@{}}
\toprule
\textbf{Work} & \textbf{AUROC} & \textbf{Accuracy} & \textbf{Sensitivity} & \textbf{Specificity} & \textbf{Precision} \\\midrule
\textbf{The study}            & \textbf{0.878 (0.859–0.897)} & 0.769 & \textbf{0.814} & \textbf{0.760} & 0.398 \\
Li et al.(2023)           & 0.794 (0.762–0.827) & \textbf{0.832} & 0.693 & 0.752 & \textbf{0.660} \\
\bottomrule
\end{tabular}
\end{table}

Our XGBoost model demonstrated a clear advancement in predictive performance, particularly in terms of AUROC, where it achieved a value of 0.878 (95\% CI: 0.859–0.897). This represents a notable 10.58\% increase compared to the AUROC of 0.794 (95\% CI: 0.762–0.827) reported by Li et al. (2023). These findings highlight the substantial improvement our model offers over existing approaches in addressing the prediction of mortality of SA-AKI patients in the ICU.

Based on the comparison of performance metrics, our model demonstrated superior sensitivity (\textbf{0.814}) and specificity (\textbf{0.760}) compared to the study by Li et al. (2023), which reported sensitivity and specificity values of \textbf{0.693} and \textbf{0.752}, respectively. Sensitivity measures the model's ability to correctly identify high-risk patients, a critical factor in ensuring timely interventions for SA-AKI patients in the ICU. Given the high stakes associated with accurately predicting patient mortality, the strong sensitivity exhibited by our model represents a critical advantage over the existing approach, ensuring a greater capacity to identify high-risk individuals effectively. The improved specificity highlights the model's precision in correctly classifying patients who are not at high risk, thereby reducing unnecessary resource allocation and false alarms.

While Li et al.'s model achieved slightly higher accuracy (\textbf{0.832}) and precision (\textbf{0.660}), these metrics come at the cost of a lower AUROC (\textbf{0.794}) and less balanced sensitivity and specificity. In contrast, our model's AUROC of \textbf{0.878 (95\% CI: 0.859–0.897)} demonstrates a substantial improvement in overall discriminatory power. This balance between sensitivity and specificity underscores the clinical utility of our approach, offering reliable predictions while maintaining practical applicability in critical care settings.




 

\section*{Discussion}

This study presents an XGBoost-driven predictive model for ICU mortality in SA-AKI patients, achieving superior performance relative to the best-reported benchmark in existing literature \cite{li2023machine}. These findings underscore the enhanced effectiveness, robustness, and precision of the model in identifying high-risk patients, highlighting its potential as a valuable tool in clinical decision-making.

\textbf{Efficient Handling of Missing Data.}
A stratified imputation strategy was employed to ensure data reliability. For numerical variables, mean imputation addressed 0–20\% missingness, KNN imputation was applied for 20–50\% missingness, and features with over 50\% missingness were excluded. Compared with traditional methods to handle with missing values, this approach minimized bias, preserved data integrity, and improved model performance.

\textbf{Comprehensive Feature Engineering.}
The breadth and refinement of feature engineering further contributed to our model’s superior performance. We began with 96 features extracted from the MIMIC-IV database, significantly surpassing the 44 features used in prior studies. This richer dataset captured more nuanced clinical details. Feature selection combined VIF, RFE, and expert input, effectively reducing multicollinearity and ensuring clinical relevance. This approach balanced dimensionality reduction and feature informativeness, yielding a robust feature set optimized for predictive accuracy.

\textbf{Rigorous Hyperparameter Optimization.}
Our model development employed GridSearchCV to systematically explore a wide range of hyperparameters, ensuring the optimal combination of settings. This exhaustive approach minimized the likelihood of suboptimal configurations, enhancing model stability and performance. Compared to prior studies, this method provided a more fine-tuned and efficient model.

\textbf{Enhanced Dataset and Updated Population.}
The updated MIMIC-IV database provided a larger study population of 9474 patients, compared to 8129 in previous research. Additionally, our stricter data extraction criteria ensured the dataset's quality and representativeness. These refinements improved generalizability while addressing potential biases inherent in earlier datasets.

\textbf{External Validation.}
External validation was conducted using the eICU Collaborative Research Database, enhancing the model's generalizability and robustness across diverse clinical settings. By validating on a dataset distinct from the development cohort, the model's performance was tested against variations in patient demographics, clinical practices, and data collection methods. This approach demonstrated the model's adaptability and reduced the risk of overfitting, confirming its reliability beyond the original MIMIC-IV dataset. Moreover, validating on eICU highlighted the model's potential for broader clinical application, supporting its use in varied ICU environments and enhancing its credibility as a decision-support tool in managing SA-AKI patients.

The superior performance of our model can be attributed to the integration of advanced missing data handling, a more comprehensive and clinically informed feature set, rigorous hyperparameter tuning, the use of a larger, high-quality dataset, and external validation. Collectively, these improvements contributed to enhanced sensitivity and specificity, ensuring more accurate identification of high-risk patients. By achieving a better balance between predictive power and clinical applicability, our model demonstrates significant advancements over existing approaches, offering a reliable tool for improving ICU SA-AKI patient outcomes.

While the study employed robust imputation strategies and feature selection methods, certain limitations should be acknowledged. The imputation techniques assumed that missing data were missing completely at random (MCAR) or missing at random (MAR), which, if violated, could introduce bias into the model. Additionally, the feature selection process, which combined statistical approaches with expert input, ensured clinical relevance but may have introduced subjective bias, potentially overlooking predictive variables. Furthermore, despite efforts to enhance interpretability through advanced selection techniques and SHAP/LIME analyses, the complexity of the XGBoost model may still pose challenges for widespread implementation in resource-limited settings or for non-technical clinical users.

Although external validation using the eICU database demonstrated the model’s robustness and generalizability, the results revealed areas for further refinement. Differences in patient populations, clinical practices, and data collection methodologies between MIMIC-IV and eICU may have affected model performance. Future work should explore adaptive learning strategies and fine-tuning techniques to optimize the model’s accuracy and utility across diverse clinical environments. These limitations underscore the need for ongoing validation and optimization to enhance the model’s reliability and clinical applicability.

\section*{Conclusion}
This study introduces a high-performing machine learning model for predicting mortality in SA-AKI patients, achieving an AUROC of 0.878 (95\% CI: 0.859–0.897), sensitivity of 0.814, and specificity of 0.760. These results demonstrate significant improvements compared to the best existing model, which reported an AUROC of 0.794 (95\% CI: 0.762–0.827). The proposed model strikes a critical balance between sensitivity and specificity, ensuring accurate identification of high-risk patients while minimizing false positives.

Key advancements contributing to this performance include a stratified strategy for handling missing data, a comprehensive feature selection process integrating statistical methods (VIF, RFE) with clinical expertise, and rigorous hyperparameter optimization via GridSearchCV. Additionally, the updated MIMIC-IV database provided a larger and more refined cohort of 9474 patients, enabling greater generalizability and reliability in predictions. External validation using the eICU database further confirmed the model’s robustness and adaptability across diverse clinical settings, highlighting its potential for broader clinical application.

While the model demonstrated strong metrics, limitations such as assumptions about missing data patterns, reliance on a single-center dataset, and the complexity of the XGBoost framework highlight areas for improvement. Future work will focus on enhancing model adaptability through advanced imputation methods, optimizing feature selection for diverse populations, and simplifying the model architecture to facilitate practical implementation in clinical settings.

By combining superior performance metrics with interpretable outputs, and validated in external database, the proposed model offers a clinically actionable tool for early identification of SA-AKI patients with high mortality risk, ultimately aiding in improving outcomes and optimizing resource allocation in critical care environments.

\bibliography{sample}

\section*{Author contributions statement}

S.C. conceived the study, designed the methodology, conducted the experiments, analyzed the data, and wrote the original draft. J.F. participated in conducting the experiments and contributed to manuscript writing. E.P., K.A., and G.P. provided critical feedback on the research methodology and manuscript. M.P. supervised the project. All authors reviewed and approved the final manuscript.

\section*{Declaration}
The data used in this study are from two publicly available critical care databases: (1) MIMIC-IV (Medical Information Mart for Intensive Care, version IV) is available through PhysioNet at: https://physionet.org/content/mimiciv/3.1/. Access requires completion of the CITI Program's "Data or Specimens Only Research" training and signed Data Use Agreement. (2) eICU Collaborative Research Database is available via PhysioNet at: https://physionet.org/content/eicu-crd/2.0/. Access requires credentialing through the eICU Institute.

Both datasets are de-identified and comply with HIPAA Safe Harbor requirements. Due to privacy protections, researchers cannot redistribute the raw data. Processed analytical datasets and code supporting the findings are available from the corresponding author upon reasonable request.

\end{document}